\documentclass[sigconf]{acmart}
\usepackage{makecell}
\usepackage{graphicx}
\usepackage{booktabs}
\usepackage{subcaption}
\usepackage{xcolor}
\usepackage{multirow}
\usepackage{url}
\usepackage{textcomp}
\usepackage[titlenumbered,ruled,noend]{algorithm2e}
\usepackage{mathtools}
\usepackage{bbold}
\usepackage{pgfplots}
\usepackage{tikz}

\DeclarePairedDelimiter\floor{\lfloor}{\rfloor}

\AtBeginDocument{%
  \providecommand\BibTeX{{%
    \normalfont B\kern-0.5em{\scshape i\kern-0.25em b}\kern-0.8em\TeX}}}

\acmConference[KDD '21]{Proceedings of the 6th Outlier Detection and Description Workshop, ACM SIGKDD Conference on Knowledge Discovery and Data Mining}{August 15, 2021}{Virtual Event, Singapore}
\setcopyright{rightsretained}
\begin{document}

% \title{Detecting Anomalies in Shipment Records Through Contrast}
% \title{Can Assumptions be Averted while Detecting Anomalies in Tabular Records ? }
\title{
Scrutinizing Shipment Records To Thwart Illegal Timber Trade}
\author{Debanjan Datta}
\affiliation{%
  \institution{Department of Computer Science, Virginia Tech}
  \city{Arlington}
  \state{Virginia}
  \country{USA}
  \postcode{22203}
}

\author{Sathappan Muthiah}
% \email{sathap1@vt.edu}
\affiliation{%
  \institution{Department of Computer Science, Virginia Tech}
  \city{Arlington}
  \state{Virginia}
  \country{USA}
  \postcode{22203}
}

\author{John Simeone}
\affiliation{%
  \institution{Simeone Consulting LLC}
  \city{}
  \state{New Hampshire}
  \country{USA}
}

\author{Amelia Meadows}
\affiliation{%
  \institution{World Wildlife Fund}
  \city{Washington, DC}
  \country{USA}
}

\author{Naren Ramakrishnan}
% \email{naren@cs.vt.edu}
\affiliation{%
  \institution{Department of Computer Science, Virginia Tech}
  \city{Arlington}
  \state{Virginia}
  \country{USA}
  \postcode{22203}
}

\renewcommand{\shortauthors}{Datta, et al.}

\begin{abstract}

Timber and forest products made from wood, like furniture, are valuable commodities, and like the global trade of many highly-valued natural resources, face challenges of corruption, fraud, and illegal harvesting.
 % Timber being a primary commodity in global commerce faces challenges of corruption, fraud, and illegal harvesting and trade.
These grey and black market activities in the wood  and forest products sector are not limited to the countries where the wood was harvested, but extend throughout the global supply chain and have been tied to illicit financial flows, like trade-based money laundering, document fraud, species mislabeling, and other illegal activities.
The task of finding such fraudulent activities using trade data, in the absence of ground truth, can be modelled as an unsupervised anomaly detection problem.
However existing approaches suffer from certain shortcomings in their applicability towards large scale trade data.
Trade data is heterogeneous, with both categorical and numerical attributes in a tabular format. 
The overall challenge lies in the complexity, volume and velocity of data, with large number of entities and lack of ground truth labels.
To mitigate these, we propose a novel unsupervised anomaly detection --- \textbf{C}ontrastive Learning based \textbf{H}eterogeneous \textbf{A}nomaly \textbf{D}etection (CHAD) that is generally applicable for large-scale heterogeneous tabular data.
We demonstrate our model CHAD performs favorably against multiple comparable baselines for public benchmark datasets, and outperforms them in the case of trade data. 
More importantly we demonstrate our approach reduces assumptions and efforts required hyperparameter tuning, which is a key challenging aspect in an unsupervised training paradigm.
Specifically, our overarching objective pertains to detecting suspicious timber shipments and patterns using Bill of Lading trade record data.
In order to target records pertinent to wood and forest products, we utilize domain knowledge curated from multiple sources. Detecting anomalous transactions in shipment records can enable further investigation by government agencies and supply chain constituents.
% Thus the proposed anomaly detection approach is suitable for such a real-world application framework.

\end{abstract}






\maketitle
\section{Introduction}
Detecting suspicious activities in international trade is a persistent challenge faced by law enforcement agencies.
Such suspicious activities can include various types of fraud, illicit financial flows, labor rights violations, unlawful wood harvesting, and violations in trade regulations.
The focus of our work is building a detection framework targeted towards identifying suspicious shipments in global trade.
The United States (US) is the largest importer of timber and wood products globally, valued at $\$51$ billion in 2017, which accounted for over $20\%$ of the global wood and timber products trade.
Illegal trade in threatened timber species  that are at risk of extinction is detrimental to not just ever-diminishing biodiversity, but also adversely impacts developing economies and is linked to national security~\cite{lawson2010illegal}.
The mislabelling (unintentional) or fraudulent (intentional) declaration of species to US consumers has been investigated ~\cite{wiedenhoeft2019fraud} and poses a significant problem, though it is unknown whether the false claims exhibited to the consumer were also present on US import declarations.

\begin{figure}[tp]
    \centering
    \includegraphics[width=0.90\columnwidth]{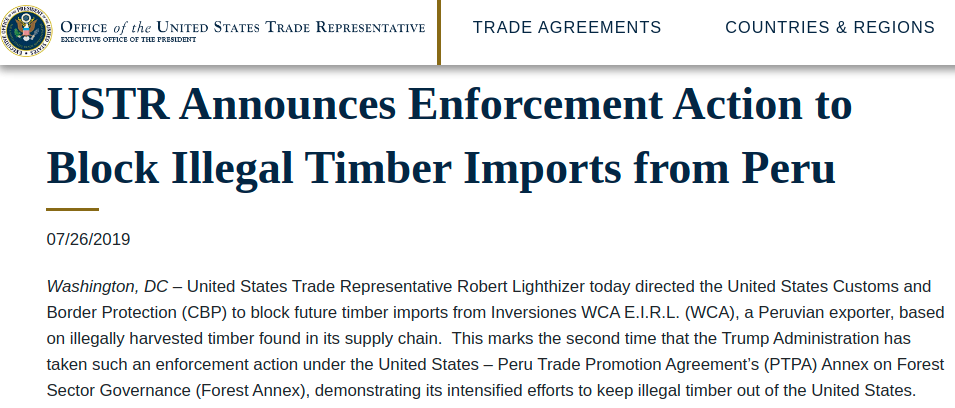}
    \caption{Enforcement remains a challenge for the initiatives and legislation enacted by the US government agencies to thwart illegal timber imports. }
    \label{fig:my_label}
\end{figure}
Finding suspicious transactions can allow agencies to target which shipments need additional scrutiny upon import into the US, and to determine patterns of suspicious trades involving specific companies and trade channels. To achieve a near real-time process that is actionable by enforcement agencies, however, requires an automated framework to detect potentially suspicious timber trades given the volume and velocity of trade data.
While manual examination and partially algorithmic tools are utilized by agencies, an integrated framework to highlight actionable trades is not readily available.
This can be attributed to factors such as hurdles in policy implementation, inter-agency communication and information-sharing agreements, handling how non-governmental organizations and government agencies share sensitive information that might lead to an investigation and prosecution, and the lack of unified domain knowledge that can be incorporated into such systems and the absence of annotated data.
% The last problem is due to prohibitively high cost of human annotation for trade data as well as sensitive nature of the trade data.
% Most importantly there are technical challenges in building such a system.

Following prior approaches towards the task of fraud detection, we formulate the task as unsupervised anomaly detection problem. 
% However the existing approaches for anomaly detection in mixed or heterogeneous data have a few drawbacks, which needs to be addressed given the nature of our data. 
Unsupervised anomaly detection in large scale heterogeneous tabular data is a non-trivial task, and existing approaches have certain drawbacks which we aim to address in this work. 
The challenge of the task  is driven by the complexity of trade data, with multiple categorical attributes with high arity (cardinality) as well as numerical attributes which contain vital information and cannot be ignored.
Secondly, model designs often have sensitive hyperparameters such as cluster numbers which are difficult to determine apriori in a true unsupervised scenario. 
Thirdly, there is often a reliance on various distributional assumptions --- which may be not be satisfied by arbitrarily complex real world data.
Thus it is imperative to overcome these impediments for our task and make the model free of such limiting assumptions, so as to create a automated suspicious shipment detection framework applicable to complex real-world trade data.

We propose a novel model, Contrastive Learning based Heterogeneous Anomaly Detector~(CHAD) that attempts to alleviate these problems.
The model uses an autoencoder based architecture to obtain a low-dimensional latent representation of data, with additional network features to handle potentially high arities in categorical variables. To estimate the density of data in this latent space we leverage the idea of Noise Contrastive Estimation~\cite{gutmann2010noise}. 
To summarize, our contributions in this work are as follows: \begin{enumerate}
    \item Anomaly detection approach based on direct likelihood estimation. Unlike prior approaches we do not make any distributional assumptions, but instead rely on contrastive learning for flexibility and generalizability. Additionally, it is robust to choice of hyperparameters.
    \item A novel and efficient negative sampling approach for heterogeneous data, that is used to train the model.
    \item Experimental evaluation demonstrating the improved performance of our model against competing approaches for trade data. Additional evaluation on multiple open source datasets show improved or competitive performance as well.
\end{enumerate}

\begin{figure}[!tp]
\centering
\includegraphics[width=\columnwidth]{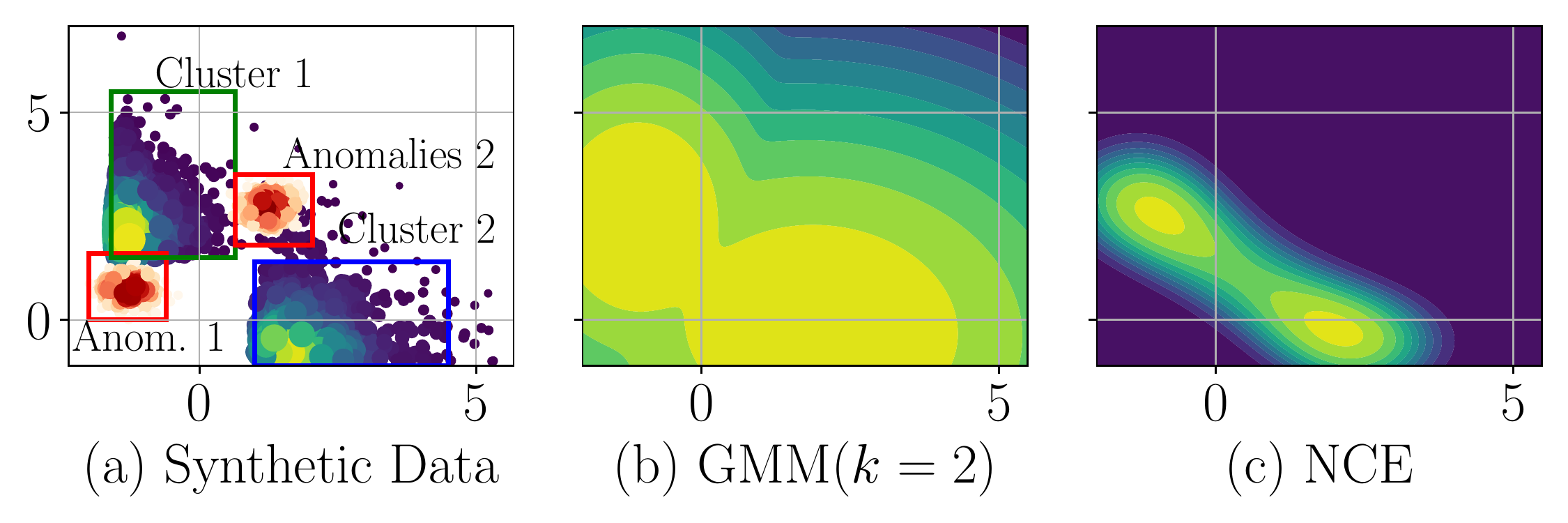}
\caption{(a) shows an example scenario where data clusters are of arbitrary shape. (b) show the density as learned by GMM and (c) the density learned through contrast using Noise Contrastive Estimation. We can clearly see that the Gaussian assumption made by GMM does not generalize well to such data and thus can significantly affect anomaly detection performance}
\label{fig:concept}
\vspace{-1em}
\end{figure}
\section{Related Works}{
% {\color{red} say we organize related work into 3 categories: 1) lazy evaluation 2) clustering based methods and 3) reconstruction based methods }
% Anomaly detection models applicable in a scenario depends on the task, nature of data as well learning paradigm.
% Anomaly detection for heterogeneous data has been explored in 
% application specific contexts such as fraud detection~\cite{abdallah2016fraud} and intrusion detection~\cite{khraisat2019survey}. \citeauthor{chalapathy2019deep}(\citeyear{chalapathy2019deep}) present a survey on the same.
% Anomaly detection methods have been surveyed extensively in \citeauthor{chalapathy2019deep}(\citeyear{chalapathy2019deep}), and explored in  
Unsupervised anomaly detection methods for general data have been discussed in detail in surveys~\cite{chandola2009anomaly, goldstein2016comparative,chalapathy2019deep}.
Application specific approaches like fraud detection~\cite{abdallah2016fraud} and intrusion detection~\cite{khraisat2019survey} have been explored in prior works.
There are prior works that focus on anomaly detection in specific types of data --- such as graphs~\cite{akoglu2015graph} and images~\cite{berg2019unsupervised}.
We limit our discussion to some of the more popular techniques and organize similar approaches together.

Traditional approaches to anomaly detection include Kernel Density Estimation~\cite{latecki2007outlier}, Principal Component Analysis and Robust PCA~\cite{xu2010robust}. 
Some other notable methods are Local Outlier Factor\cite{breunig2000lof} and  Isolation Forest~\cite{liu2008isolation}.
A key challenge with such methods is that they are not scalable, and many of them are not effective for sparse high dimensional input features.
% ====================
% Clustering
% ====================

\textbf{Clustering} based and geometric anomaly detection approaches assume \textit{normal} or expected data to posses proximity to each other along with some latent underlying structure, which has been can be utilized to detect outliers. 
Recent approaches like Deep Embedded Clustering (DEC)~\cite{xie2016unsupervised} and Deep Clustering Network (DCN)~\cite{yang2017towards} are relevant here.
Mixture models based approaches like 
DAGMM~\cite{zong2018deep} which combines a deep autoencoder with a Gaussian mixture model to perform anomaly detection can also be considered under clustering methods.
An interesting aspect of DAGMM is a very low dimensional latent representation is augmented with sample reconstruction errors.

\textbf{Autoencoders} utilize a reconstruction-based approach relying on the assumption that anomalous data cannot be represented and reconstructed accurately by a model trained on \textit{normal} data.
Autoencoders and its variants such as denoising autoencoders have been used for anomaly detection~\cite{sakurada2014ae,tagawa2015structured,zhou2017anomaly}, along with ensemble based approaches~\cite{chen2017outlier}.

\textbf{One-class classification} based approaches like OCSVM~\cite{scholkopf2000support} separates the normal data from the anomalies using a hyper-plane of maximal distance from the origin.
Support Vector Data Description~\cite{tax2004support} attempts to find the smallest hyper-sphere that contains all normal data. 
Both methods use a hyperparamter $\nu$ that helps define the boundary of the hyper-plane or hyper-sphere. 
Similarly, two recent works employ deep learning along with a one-class objective for anomaly detection.
DeepSVDD~\cite{ruff2018deep} performs anomaly detection combining deep convolutional network  as feature extractor and one-class classification based objective. 
Oneclass-NN~\cite{chalapathy2018anomaly} uses a similar formulation, albeit using a hyper-plane instead of a hyper-sphere, combined with an alternating minimization based training approach.

%--------------------------
% Categorical Data
% -------------------------
Methods for anomaly detection in purely multivariate categorical data include information theoretic methods such as 
\textit{CompreX}~\cite{akoglu2012fast}, and embedding based methods such as \textit{APE}~\cite{chen2016entity} and \textit{MEAD}~\cite{datta2020detecting}. 
Though \textit{MEAD} uses a trade dataset, it considers only categorical features.
We do not consider these methods for comparison since discretization of continuous variables require careful considerations and can lead to information loss~\cite{virkar2014power}.
% Both represent entities in a common embedding space and interactions between them are computed to obtain the likelihood of a record. 
% -------------------------
Many of the above mentioned approaches make critical assumptions about the data such as cluster numbers or distributional family, etc. 
These are often specific to a particular kind of data and also it is difficult to obtain necessary domain knowledge to be able to make such choices.
In the following section we present a motivating example demonstrating the disadvantages of making such assumptions about data and propose an alternative non-parametric density estimation method.
}

\section{Data and Motivation}
In this section we briefly outline the constraints, requirements and motivation for the data driven system and provide further background on the context of the application. We also present the conceptual motivation in proposing our anomaly detection model for the task, outlining the applicability of such an approach for real world data.

\subsection{Trade Data}
% The shipment data utilized for experimental evaluation is the real Bill of Lading records from U.S. import data.
High resolution trade data contain heterogeneous features with high arity categorical features and numerical features.
We specifically use US import Bill of Lading records for our framework from 2015 to 2017 obtained from Panjiva~\cite{panjiva-trade-data-2019}.
The raw data is preprocessed, since there are many attributes of which a subset are deemed irrelevant. Additionally many attributes have large percentages of missing data. 
With the help of domain experts we select the following attributes :\textit{ VolumeTEU}, \textit{Quantity}, \textit{WeightKg}, \textit{NumberOfContainers}, \textit{Carrier}, \textit{HSCode}, \textit{PortOfLading}, \textit{PortOfUnlading}, \textit{ShipmentDestination} and \textit{ShipmentOrigin}.
The first four are real valued attributes, and the remaining are categorical.

Goods and products involved in global trade are tracked using the standardized Harmonized Schedule (HS) code nomenclature and product classification system. The role of HS codes is important in the context of trade data.
We obtain the ontology and data for HS codes from open source repositories containing text descriptions of products, which can include scientific names, family and common names of endangered and commercially viable timber species. 
We use natural language processing for parsing and lemmatizing, and use regular expression and n-gram based matching to obtain these from the HS code text descriptions after collating HS code data from multiple years.  
With the help of domain experts, we extract HS codes that can be associated with known high risk timber species.
Additionally, we curate HS codes covered by legislation such as the Lacey Act's Plant Declaration Form and data on country-specific logging and export bans.
HS codes for products containing solid wood (like furniture) are used to filter out trade records for our system.
Although these curated HS codes may contain high risk species, they correspond to such a large number shipments that simple rule-set based matching is neither analyzable nor actionable by end users.

\subsection{Conceptual Motivation}\label{sec:concept}
Geometric approaches anomaly detection are based on the assumption that nominal data points are clustered, in the original or a transformed space which is supported by the manifold hypothesis~\cite{goodfellow2016deep}. 
Assumptions are often placed on the shape of the cluster,such as hypersphere or ellipsoid for computational efficiency which the true data distributions may not conform to. Data in latent space can have complex non-linear or non-convex decision boundaries. 

We demonstrate the limitations of such assumptions using a simple scenario with data $\in$ $\mathbb{R}^2$ as shown in Figure~\ref{fig:concept}.
Two clusters ($C_1$ and $C_2)$ of nominal data points points are generated using two independent bi-variate Gamma distributions, $N_1$ and $N_2$, which are almost triangular in shape.
Another two sets of points, which are treated as anomalies are generated using Normal distributions denoted as $a_1$ and $a_2$.
Let us model the data using Gaussian Mixture Model (GMM) with $k=2$ components and also with K-Means with $k=2$ .
We estimate the anomalousness of a point as the distance from assigned cluster center for the K-Means model and as the likelihood of a data point conditioned on the estimated model parameters for GMM (similar to DAGMM ~\cite{zong2018deep}). 
For the K-Means based model and the GMM based model, we find the average precision(auPR)  to be $<0.5$ and $\approx 0.8$ respectively.
Here $k$ is chosen as $2$ for the K-means based model and shows a reasonably good performance. 
However such knowledge is not readily available in complex real-world data. The average precision drops significantly to $0.2$ when changing the number of clusters to $k=1$. 
A single cluster assumption is not invalid, and made in cases like SVDD~\cite{tax2004support} and deepSVDD~\cite{ruff2018deep}.
While methods have been proposed to estimate the optimal number of clusters for a given dataset such as X-means~\cite{pelleg2000x} or Bayesian GMM~\cite{rasmussen2000infinite}, they are known to suffer from scalability issues in complex (high dimensional) and big data models. 
While for a simple scenario like the one considered above, determining the appropriate number of clusters through tests such as silhouette analysis would be easy --- but for sparse high dimensional data such analysis is prone to errors.
We can depart from such limiting assumptions if we could approximate a \textit{contrastive} distribution, whose density is high only in regions where the \textit{normal} data is absent. 
Provided $G_1$ and $G_2$ is revealed to us, this can be easily done by sampling from an uniform distribution $U$ over the valid data domain such that $P_{G_1}(u)< \epsilon$ and $P_{G_2}(u)< \epsilon$. 
% Here $u \in U$, $\epsilon$ is a small threshold value and we assume that $G_1$ and $G_2$ is revealed to us.
Sampling from this distribution, and from the \textit{normal} data, it is possible to create a discriminator function that can predict which region of space a point belongs, or in other words can estimate the probability of a point belonging to region of \textit{normal} data following the concept of NCE~\cite{gutmann2010noise}. Such a model achieves an average precision $\approx 0.9$. 

This shows a model that estimates the density of \textit{normal} or expected data using a contrastive distribution can be effectively used for anomaly detection. 
% A visual metaphor might be looking the \textit{negative} of a photograph which tells us in which areas there are objects of interest and which areas are void.
The key challenge is that in real world data neither is analytical form of data distribution exactly known, nor is it a trivial task to estimate such a contrastive distribution. 
In the upcoming sections we detail how to overcome this problem and estimate the contrastive distribution through our model.

\section{Model Architecture}\label{sec:model}
\begin{figure}[t]
\centering
\includegraphics[width=0.80\columnwidth]{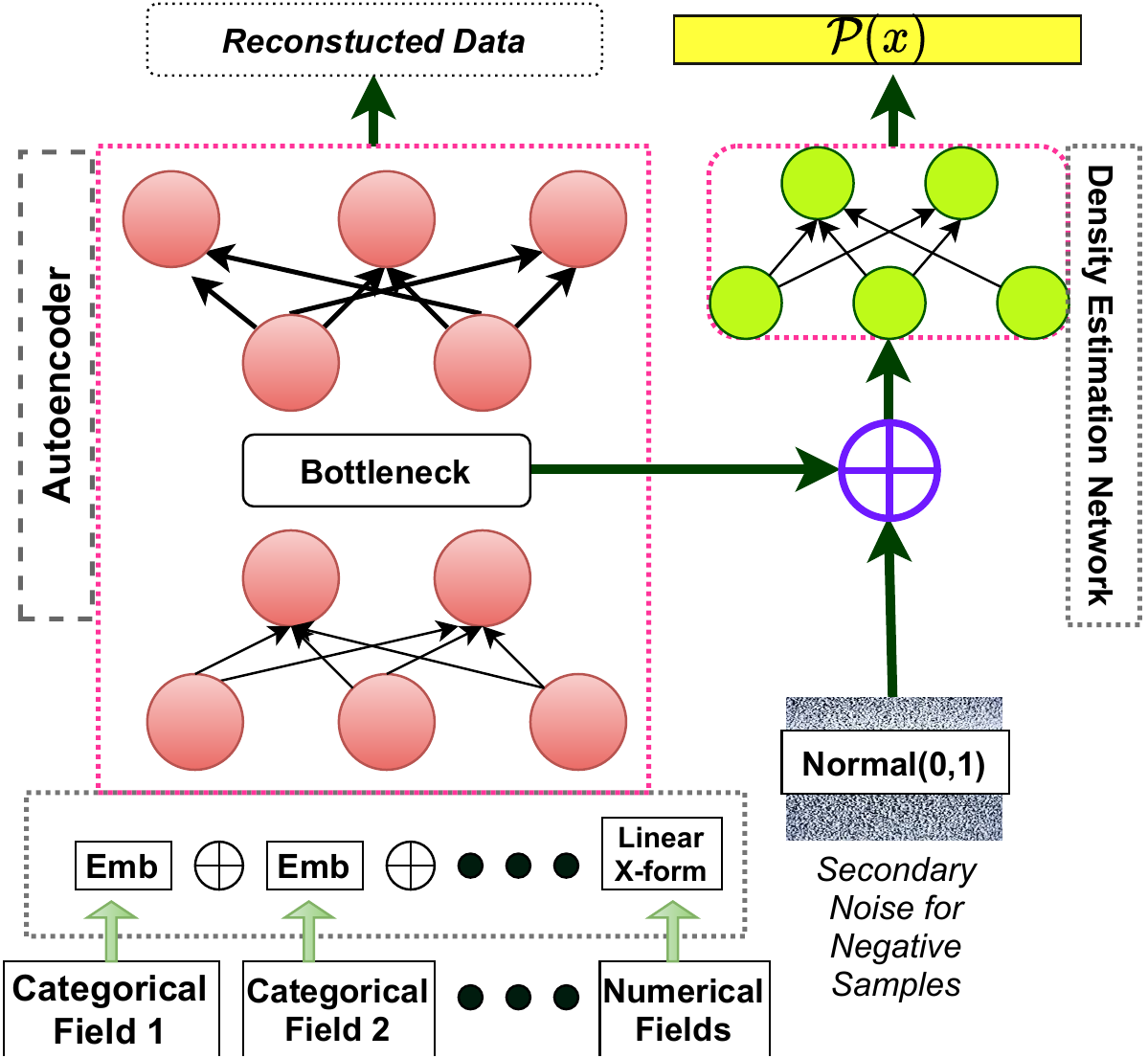}
\caption{Proposed architecture of our model Contrastive Learning based Heterogeneous Anomaly Detector(CHAD)}
\label{fig:model}
\end{figure}

%=================\
\textbf{Problem Statement}: Given a dataset $\mathcal{D}$ containing heterogeneous attributes which are assumed to be clean, learn a model $\mathcal{M}(\theta)$ that can predict $P(x |\mathcal{\theta})$, the likelihood of a test record being drawn from the underlying data distribution. 
The test records with likelihood below a user-defined threshold are deemed anomalous.

While embedding based architectures have been used in prior works, deep autoencoders preserve richer information in a reduced dimension compared to shallow or linear autoencoders~\cite{hinton2006reducing}.
% However autoencoder based approaches presented in prior works have a drawback since they use reconstruction error as anomaly scores. 
% This is because if the autoencoder generalizes too well - it can lead to anomalous samples having low reconstruction error~\cite{spigler2019denoising}.
Use of reconstruction error as anomaly score has the associated caveat that if the autoencoder generalizes too well - it can lead to anomalous samples having low reconstruction error~\cite{spigler2019denoising}.
% In line with these observations we make use of autoencoders for modeling our data but instead of using reconstruction loss as anomalousness score we use the learned latent representation to build a classifier that learns to discriminate between observed data and random noise.
Thus while we use an autoencoder to capture a reduced dimensional representation for the data, but build a density estimation network to directly estimate the likelihood of data instances
% The classifier softmax score is then used as the score in our model. Details of our autoencoder architecture is presented next.

% =======
\subsection{Asymmetric Autoencoder}{
% Autoencoders generally have symmetric architectures for the encoder and the decoder, however we propose the use of an asymmetric architecture.
We propose the use of an asymmetric architecture that is aware of the fields in the tabular input data.
The decoder is a fully connected dense neural network with dropout.
The first fully connected layer in the encoder is concatenation of embedding(linear transform) for each categorical attribute(domain), and either linear or identity transform for the set of all continuous features. 
This decision to add the transformation depends on the dimensionality, and we find it helps to add a transform where dimensionality is over 32.

Formally, let $\mathbf{x} \in \mathcal{D}$ be an input. $\mathbf{x}$ is a multidimensional variable, with $d$ features.
Let $x_1,x_2...x_k$ be categorical features, and $x_{k+1}...x_d$ be continuous features.
$x_{k+1}...x_d$ can be considered as a multivariate feature, denoted as $x_r \in R^{d-k}$. 
Let $f_{i}(.)$ denote the linear transform or the identity transform, for $i^{th}$ categorical field. 
Let $g(.)$ denote transformation on $x_r$. Let $x_t$ denote the transformed input, which is input to the fully connected layers of the encoder, as shown in Eq.~\ref{eq:x_e}.
Both the encoder and decoder utilize dropout and a non-linear activation function(tanh), except for the final output where sigmoid is used.
\begin{equation}
   x_t = f_1(x_1) \oplus f_2(x_2) ... \oplus f_k(x_k) \oplus g(x_r)
   \label{eq:x_e}
\end{equation}
% ============== New paragraph ============= %
The autoencoder is designed to optimize reconstruction of the input vector, which is an auxiliary task in our case since we are interested in the latent representation. 
We choose Mean Squared Error for the reconstruction Loss, denoted as $\mathcal{L_R}$ as shown in Eq.~\ref{eq:loss1}. 
\begin{equation}
    \mathcal{L_R} = \frac{1}{N}\sum_{i}(\mathbf{x}-\mathbf{\hat{x}})^2
    \label{eq:loss1}
\end{equation}

While it is possible to have separate losses for categorical and numerical features, we do not find any significant performance difference in doing so.
In the next section, we detail out how the latent representation is used in the discriminator module to identify data density.
}

\subsection{Density Estimation Network}{
% We started out with the premise of estimating the probability distribution directly.
The central idea here is to learn the density of data through comparison or contrast with an artificially generated noise distribution, extending the concept of NCE~\cite{gutmann2010noise}.
Let the class labels for the normal data and noise be $1$ and $0$ respectively. 
We start with an assumption that $P(C=1)=P(C=0)=0.5$ and  denote the empirical distribution of noise as $p_n(.)$. 
We model the density of data as $p_d(.;\theta)$ where $\theta$ are the model parameters. 
That is, $P(x|C=1,\theta)=p_d(x;\theta)$ and $P(x|C=0)=p_n(x)$. 
Therefore,
\begin{equation}
    P(C=1|x;\theta) = \frac{p_d(x;\theta)}{p_d(x;\theta) + p_n(x)} = f(x;\theta)
\end{equation}
Here $P(C=1|x;\theta)$ is the posterior distribution of class label 1, and $f(x;\theta)$ is the approximate estimation function.
A MLP can be utilized as functional approximator for $f(x;\theta)$ without loss of generality. 
Our objective is to find the set of parameters $\hat{\theta}$ for the model that maximize the likelihood of the training data. 
\begin{equation}
\small
\begin{aligned}
    \mathcal{L}(\hat{\theta}) =   \arg \underset{\theta} \max \sum_{i} C_i ln(P(C_i=1|x_i;\theta) \\ + (1-C_i) ln(P(C_i=0|x_i;\theta)
    \\
    \approx \sum_{i}ln(f(x_i;\theta)) + ln\big(1 - \frac{1}{|k|}\sum_{k}(f(z_k;\theta)\big)
\end{aligned}
\label{eq:est1}
\end{equation}   
In the second line of Eq.~\ref{eq:est1}, $z_k$ refers to the $k$th negative sample drawn for each observed instance $x_i$. This is because we do not have access to actual samples from class 0.
% We find the objective to be same as Binary Cross Entropy loss.
% with signs reversed.

Now, we can replace $x_i$ with $x_e$ in Eq.~\ref{eq:est1}, where $x_e = \mathbf{f_{enc}}(x_i)$ and $x_i \in \mathcal{D}$. 
Here $\mathbf{f_{enc}}$ is the transformation function of the encoder in the autoencoder.
Specifically, let $\mathbf{f_{enc}}():R^d\mapsto R^p$, such that p is dimensionality of latent vector.
Also, for each $x_i$, we draw $K$ negative samples from a noise distribution $\mathcal{\hat{D}}$, denoted as $z_i \in R^d$.
Then, $z_e^i=\mathbf{f_{enc}}(z_i)$ with $z_e\in R^p$.
We inject secondary noise  $n$ to latent representation $z_e$ of each negative sample drawn from a multivariate Normal distribution, $n \sim N(0, \mathbf{I}^p)$, where ${I}^p$ is a identity covariance matrix. The noise injection is done to increase variation in negative samples and improve model performance. 
Simplifying the above, we have the following.
\begin{equation}
\mathcal{L}_{est}(\hat{\theta}) = 
-\gamma  \sum_{x_i \in |\mathcal{D}|}ln(f(x_e)) 
- ln\big(1 - \frac{1}{|k|}\sum_{k}(f(z_e^i+n_k))\big)
\label{eq:est2}
\end{equation}  
% We can also see that it bears resemblance with the objective function used in \cite{tang2015line}.
Minimizing this loss we can learn the parameters to capture the distribution of normal data. 
Here $\gamma$ is a penalty if the model assigns low likelihood to normal(nominal) data instances.
It is initially set to 1 and slowly increased to a maximum value in the third training phase. Empirically we find that results are not sensitive to $\gamma$. 

We utilize a simple two layered MLP with dropout to estimate $f(;\theta)$. It is found to work well in practice, though a more sophisticated estimator network can be designed.
The overall loss function to optimize, from Eq.~\ref{eq:loss1} and~\ref{eq:est2} is as follows.
\begin{equation}
    \mathbf{Loss} = \lambda \mathbb{1}(\mathbf{t}_r)\mathcal{L_R}   + \mathbb{1}(\mathbf{t}_e)\mathcal{L}_{est}(\hat{\theta})
    \label{eq:joint_loss}
\end{equation}
Here $\mathbb{1}(\mathbf{})$ is an indicator variable $\in \{0,1\}$.
We explain in Section~\ref{sec:training} how $\mathbb{1}(\mathbf{t_r})$ and $\mathbb{1}(\mathbf{t_e})$ are set depending on the training phase.
The training hyperparameter $\lambda$ modulates the importance of $\mathcal{L_R}$, and how it is varied is discussed in the training procedure. 
The overall architecture of our model, \textbf{C}ontrastive Learning based \textbf{H}eterogeneous \textbf{A}nomaly \textbf{D}etetctor is shown in Figure~\ref{fig:model}.
}

\subsection{Model Training}\label{sec:training}
We make use of a three phase training procedure to learn our model.
The first phase is termed as burn-in phrase,  is training the autoencoder only using reconstruction loss.
In this phase the estimator network is not modified.
This is essential because we want to first obtain the appropriate latent representation for the data, such that normal(nominal) data can be correctly reconstructed. 
Referring to Equation~\ref{eq:joint_loss},  $\mathbb{1}(\mathbf{t}_r)=1$ and $\mathbb{1}(\mathbf{t}_e)=0$ in this phase.
The second phase trains both the autoencoder and the estimator jointly.
We notice that the scale of the two losses can be different, we adopt an approach similar to some GANs~\cite{gulrajani2017improved}, including the estimator loss only for alternate mini-batches.
% and it is difficult to determine best weight values in an unsupervised setting.
% Inspired by the training procedure some GANs~\citeyear{gulrajani2017improved}, the estimator loss is included in alternate mini-batches.
That is $\mathbb{1}(\mathbf{t}_e)=1$ for alternate batches, $\mathbb{1}(\mathbf{t}_r)=1$ being constant. 
This leads to a stable joint training of the two components, avoiding requirement for a sensitive user provided scaling hyperparameter.
Referring to Eq.~\ref{eq:joint_loss}, the scaling hyperparameter $\lambda$ is set to 1 in first phase.In the second phase it is decayed using $exp(-t)$ where $t$ is the epoch number of the second phase. 
The third and final phase keeps the encoder parameters fixed.
The density estimation network is trained based on the 
estimator loss, with   $\mathbb{1}(\mathbf{t}_e)=1$ and $\mathbb{1}(\mathbf{t}_r)=0$.

\section{Generating Negative Samples}
Generating negative samples is key in contrastive estimation of the data distribution.
In the case of heterogeneous data, what can be construed as negative samples is not immediately evident as in other cases such as text or networks.
% However there are a couple of key criteria to satisfy.
Firstly, negative samples should have adequate variation and not be clustered such that they provide a contrastive background to estimate the data distribution.
Secondly, they should not be entirely comprised of noise or be divergent from target distribution, but have some similarity with the normal data -- so that negative samples are located within the domain of the nominal data in normal space.
% We propose a random subspace based approach applied to both the categorical and the real valued features in data. 
% For text negative samples can be simply defined as words that do not co-occur with a target word; whereas for networks negative samples are non-adjacent nodes.
% First let us consider the subset of categorical features. 
% A record where a combination of entities that does not occur in training data might be considered as a negative sample.
% This can be achieved through random perturbation of a pair of entities training record.

While unexpected combination of entities or values might be considered as negative samples, simply perturbing only the categorical features or only the continuous features might not guarantee the resulting(perturbed) record satisfying the aforementioned criteria. 
This is because we are unaware of the importance of and interactions among individual features that defines the behavior of data in the latent space.
Further, we found merely adding isotropic noise to the  latent representation of data does not provide good negative samples. 
Thus we propose the random subspace based approach outlined in Algorithm~\ref{alg:neg_sam} for obtaining negative samples.

In this approach at most half of categorical features are selected, and each entity is replaced by another instance belonging to same category.
Let $a_w$ be the arity of the $w^{th}$ category. The probability of a category to be selected for perturbation is chosen following a multinomial distribution. 
The probability of $w$ being selected is the sum normalized value of $q_w$, where $q_w = (a_w/ \sum_{w`}a_{w`})^{0.75}$. 
The dampening factor $0.75$ is added following ~\cite{mikolov2013distributed}, so that not only the high arity features are perturbed.

Continuous values are assumed to be normalized to be in the range $[0,1]$.
We add uniformly generated random noise to randomly selected $\floor{r/2}$ features, where $r$ is the number of continuous features. 
The noise deviation parameter $\delta$ is used to shift the mean of $U(0,1)$. Intuitively using $\delta=0.5$ makes sense, and works well in our case.
The range of the noise is set up to be beyond the expected range of the features.
This per-feature perturbation forces the feature values for negative samples to spread out over and beyond the range of possible values.

\begin{algorithm}[tp]
% \small
\KwData{Training records $\mathcal{D}$; noise deviation $\delta$}
\KwResult{Negative samples for each training record}
\For{each training record $\mathbf{x} \in \mathcal{D}$}{
\For{i = 1:m negative samples} {
    Select $i \in [1,\floor{k/2})$ categorical features\;
    \For{each feature in $i$}{
    Replace entity with a random entity\;
    }
    Select $|j_1| = \floor{r/4}$ real features randomly\;
    \For{each feature $u \in j_1$}{
    $n \sim$ Uniform(0,1) + $\delta$, $u = u + n $
    }
    Select $|j_2| = \floor{r/4}$ real features randomly\;
    \For{each feature $v \in j_2$}{
    $n \sim$ Uniform(0,1) - $\delta$, $v = v + n $
    }
  }
}
\caption{Negative sample generation}
\label{alg:neg_sam}
\end{algorithm}

\begin{table*}[!th]
    \centering
    \begin{tabular}
    {lccccccc}
    \toprule
    \textbf{Data Set} & \textbf{OCSVM}   & \textbf{FAE-r}         & \textbf{DAE} & \textbf{DCN} & \textbf{DAGMM} & \textbf{dSVDD} &\textbf{CHAD}\\ 
    \midrule
    KDDCup99    &  \makecell{\textbf{0.99039} \\($\pm$ 0.0012)} & \makecell{0.96113 \\ ($\pm$ 0.0371)} &  \makecell{0.76297 \\($\pm$ 0.1076)} & \makecell{0.89324 \\($\pm$ 0.0447)} & \makecell{ 0.79432  \\($\pm$  0.2015)} & \makecell{  0.68164 \\ ($\pm$0.1407) } & \makecell{ 0.97332\\ ($\pm$ 0.0158)}\\
    \midrule
    KDDCup99-N  & \makecell{0.99999 \\($\pm$ 0.0001)} & \makecell{0.58391 \\ ($\pm$ 0.0002)} &  \makecell{0.99957 \\($\pm$ 0.0004)} & \makecell{0.94701 \\($\pm$ 0.0312)} & \makecell{0.99695 \\($\pm$ 0.0025)} & \makecell{ 0.97843 \\($\pm$ 0.0046)} & \makecell{ 0.99647\\ ($\pm$ 0.0035)}  \\
    \midrule
    UNSW-NB15   & \makecell{0.38445 \\($\pm$ 0.0055)} & \makecell{0.65070 \\ ($\pm$ 0.0379)} & \makecell{0.26121 \\($\pm$ 0.0498)} & \makecell{0.66786 \\($\pm$ 0.0785)} & \makecell{0.34278 \\($\pm$ 0.1980 )} & \makecell{ 0.74479 \\($\pm$ 0.1144)} & \makecell{\textbf{0.78873} \\ ($\pm$ 0.0032)}  \\
    \midrule
    NSL-KDD     & \makecell{0.84869 \\($\pm$ 0.0059)} & \makecell{\textbf{0.88895} \\ ($\pm$ 0.0316)} & \makecell{0.81286 \\($\pm$ 0.0329)} & \makecell{0.53152 \\($\pm$ 0.1574)} & \makecell{0.36002 \\($\pm$ 0.1440)} & \makecell{  0.60989 \\($\pm$ 0.0729)} & \makecell{ 0.75512 \\ ($\pm$ 0.0517)}  \\
    \midrule
    Gure-KDD    & \makecell{0.68652 \\($\pm$ 0.0063)} & \makecell{0.67999 \\ ($\pm$ 0.0229)} &  \makecell{0.68553 \\($\pm$ 0.0433)} & \makecell{0.37014 \\($\pm$ 0.0981)} & \makecell{0.35806 \\($\pm$ 0.0834)} & \makecell{  0.63064 \\($\pm$ 0.0719)} &\makecell{ \textbf{0.75680} \\ ($\pm$ 0.03918)}  \\
    \midrule
    \textbf{Average Score} & 0.78201 & 0.75294 & 0.70443  & 0.68195 & 0.56960 & 0.72908 & \textbf{0.85409}\\
    \bottomrule
    \end{tabular}
    \caption{Performance comparison w.r.t to average precision (auPR). On average CHAD outperforms all models with an improvement of over 9.2\%  over second best model. Though CHAD is not the best in KDDCup99 and NSL-KDD datasets it does provide competitive results. For KDDCup99-N, we find there is no single definitive best performing model. Experimentally we find that unlike CHAD the other models are highly sensitive to choice of hyper-parameters (refer Fig.~\ref{fig:hyperparams}) and in absence of labelled data /domain knowledge making a good choice for hyper-parameters will be difficult. }
    \label{tab:table_results}
\end{table*}

\section{Model Evaluation}
% In this section we provide details on the experiments performed to evaluate the efficacy of our model. We perform comparative evaluation for publicly available and our trade datasets.
% =======================================

\subsection{Baselines for Comparison}
The metric used for evaluation performance of our model against competing baselines is chosen as \textit{average precision}, which is the area under the precision-recall curve following prior works~\cite{akoglu2012fast,davis2006relationship}
The following baselines are chosen for comparing model performance.

\textbf {OCSVM}~\cite{scholkopf2000support} is a classification based approach, that determines a decision boundary for normal data. 
An exponential kernel is used and hyperparameter $\nu$ is set to a standard value.

\textbf{Deep SVDD}~\cite{ruff2018deep} proposed deep learning based feature extraction combined with a one class classification ~\cite{tax2004support}, ~\cite{scholkopf2000support}, with two variations of the objective --- one-class and soft-boundary. 
We replace the CNN based feature extractor presented in the original work with a three layered DNN with dropout for general data. 
% Anomaly scores are determined as distance from the centre of the hyper-sphere, defined by points in latent space, 

\textbf{DAGMM}~\cite{zong2018deep} combines a deep autoencoder with Gaussian mixture model to perform anomaly detection.
It augments the latent data representation with reconstruction losses, and clusters the points into mixture of multivariate Gaussian distributions.
The estimated likelihood is used as anomaly score for the samples.
% The suggested 3 layered network architecture and other parameters for the KDDCup99 data set in original work is used.
% , and similar architectures are used for the remaining data sets. 
% We run the model with variations of with latent-representation size and number of clusters and report best results.

\textbf{DCN}~\cite{yang2017towards} performs dimensionality reduction through a deep autoencoder and K-means clustering jointly. 
We use greedy layer-wise pretraining for the autoencoder to improve performance, prior to performing joint optimization. 
The distance of a point from its assigned cluster center is used as the anomaly score.
% We use architectures similar to DAGMM  and tune for the number of clusters and other training parameters. 

\textbf {Deep Auto Encoder}~\cite{chen2017outlier}: A deep autoencoder with dropout with greedy layer-wise pretraining is used, and reconstruction loss is used as anomaly score.
% Greedy layer-wise pretraining is used, along with dropout to train the model. 
Previous works~\cite{xie2016unsupervised} point out that successive layers of dropout achieve a denoising effect. 
% The network architecture used is similar to DCN.
% with some changes for tuning performance.

\textbf {FAE-r}: We use field aware auto-encoder from our CHAD with reconstruction loss as anomaly score, without any modification such as greedy layer-wise pretraining.
\vspace{-0.5em}
\begin{figure*}[tp]
    \centering
    \includegraphics[width=\textwidth]{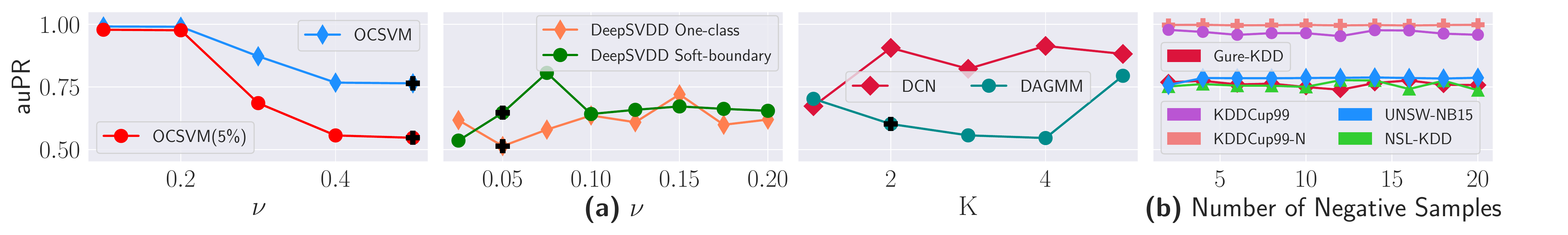}
     \caption{(a) Performance variation of competing methods to changing key hyper-parameters, for anomaly detection on KDDCup99 data. The standard or recommended hyperparameter settings are highlighted. (b) Effect of varying number of negative samples per instance on model performance for CHAD for all datasets.}
     \label{fig:hyperparams}
\end{figure*}
\subsection{Experimental Results on Intrusion Detection Data}
To better understand the performance of our model on we choose open source intrusion detection datasets, which are as follows:
% \begin{enumerate}
%     \item 

\textbf{KDDCup99}~\cite{Dua:2019}:
The KDDCup99 10 percent dataset has been a benchmark dataset for anomaly detection tasks. 
We choose the 'Normal' class label as normal data and attack classes are treated as anomalies.\\
% \item  
\textbf{KDDCup99-N}~\cite{Dua:2019}: 
We consider the instances KDDCup99 10 percent data with attack class `Neptune' to be normal data, and data points with label 'Normal' are considered as be anomalies. \\
% \item  
\textbf{UNSW-NB15}~\cite{unswnb15}: 
% This dataset contains both real and synthetic network data.
We select data with label \textit{Normal} as normal data, and classes \textit{Backdoor}, \textit{Analysis}, \textit{Shellcode}, \textit{Worms} as anomalies. \\
% \item  
\textbf{NSL-KDD}~\cite{nslkdd_dataset}: 
The combined train and test partitions of the original dataset is used, with 
`Normal' class chosen as normal data and rest as anomalies.\\
% \item 
\textbf{GureKDD}~\cite{gureKDD_dataset}: The `Normal' class label is chosen as normal data, and rest as anomalies.\\
% \end{enumerate}
For all models, we experiment with various choices for the respective hyper parameters and report the best score. 
We refer the reader to Section~\ref{subsec:hyperparams} for more details.
The experimental results are shown in Table~\ref{tab:table_results}. 
We follow the same experimental protocols followed for the trade data. 
For our model CHAD, we use 10 negative samples for training the density estimation network.
We use Adam~\cite{kingma2014adam} for optimization for all methods that require it, with learning rate to $5 \times 10^{-3}$ and batch size 256 or 512 across all datasets. 
The number of epochs for the three training phases are set to 50,10 and 25 respectively. We use a generic 3-layered pyramid-like architecture for the autoencoder, with a dropout of 0.2. 

In judging average performance we find our model CHAD performs better than the baselines. 
% For the baselines, we run multiple parameter settings, and report the best results.
OCSVM provides comparable performance in some cases.
% , being a one-class classification based approach. 
However, it does not perform well for UNSW-NB15 dataset which can be attributed to either nature of the data and higher dimensionality caused by comparatively larger entity count.
% higher arity of a categorical variable.
DAE and DCN also performs well in most cases. 
DAGMM however does not perform well despite its expected good performance on general data. We observe a high variance, and it can be attributed to the fact that loss metrics augmenting the latent space does not necessarily improve model expressiveness.
FAE-r is shown to perform well in many cases, which demonstrates that our autoencoder captures the latent representation well.
However, it performs poorly on KDDCup-N, which can be attributed to over generalization issue mentioned earlier.
We find our model performs comparably or favorably in majority of cases. 
% For KDDCup99-N, most models perform very well, alluding to it is a comparitively less challenging task to find anomalies in this dataset.
Our model does not perform the best for NSL-KDD, however in case of UNSW-NB15 and Gure-KDD it shows a strong advantage. However it is to be noted for baselines such OCSVM, DAGMM and DCN  we experiment with different hyper-parameters using a holdout of test samples, which is not possible in real-world scenarios.
In Section~\ref{subsec:hyperparams} we further  examine and demonstrate how  competing models are sensitive to bad choices of hyper-parameters.
\begin{table}[t]
\small
\centering
\begin{tabular}{c|cccc}
\toprule
\makecell{Trade \\ Dataset}  & \makecell{Total\\cardinality}&  \makecell{Numeric\\ features} & Train size & Test size \\
\midrule
Dataset-1 & 809  & 4  & 90910   & 57772   \\
Dataset-2 & 853  & 4  & 106858  & 57213  \\
Dataset-3 & 850  & 4  & 96545   & 51189   \\
Dataset-4 & 859  & 4  & 108841  & 56298  \\
Dataset-5 & 844  & 4  & 109525  & 50481  \\
Dataset-6 & 877  & 4  & 113692  & 56432  \\
\bottomrule
\end{tabular}
\caption{Details of the trade datasets used for experimental evaluation. Total cardinality here refers to the sum of the cardinality of the categorical attributes.}
\label{table:trade_data}
\end{table}

\begin{table}[!tp]
\centering
\begin{tabular}
{ l|cccc}
\toprule
\textbf{Data Set} & \textbf{OCSVM} & \textbf{FAE-r} & \textbf{dSVDD} & \textbf{CHAD}\\ 
\midrule
Dataset-1 & \makecell{0.931139 \\ ($\pm$ 0.0003)} &  \makecell{0.562055 \\ ($\pm$ 0.0169)} &\makecell{0.630166 \\ ($\pm$ 0.0964)} & \makecell{\textbf{0.982815} \\ ($\pm$ 0.0053)}  \\      
\midrule
Dataset-2 & \makecell{0.170002 \\ ($\pm$ 0.0400)} & \makecell{0.577034 \\ ($\pm$ 0.0149)} &\makecell{0.536559 \\ ($\pm$ 0.0652)}  & \makecell{\textbf{0.996048} \\ ($\pm$ 0.0010)} \\   
\midrule
Dataset-3 & \makecell{0.943908 \\ ($\pm$ 0.0001)} & \makecell{0.597657 \\ ($\pm$ 0.0159)} &\makecell{0.681765 \\ ($\pm$ 0.0839)} & \makecell{\textbf{0.992471} \\ ($\pm$ 0.0019)} \\ 
\midrule
Dataset-4 & \makecell{0.997622 \\ ($\pm$ 0.0001)} & \makecell{0.989836 \\ ($\pm$ 0.0031)} &\makecell{0.755632 \\ ($\pm$ 0.0209)} & \makecell{\textbf{0.999369} \\ ($\pm$ 0.0002)} \\  
\midrule
Dataset-5 & \makecell{0.319723 \\ ($\pm$ 0.0010)} & \makecell{0.939353 \\ ($\pm$ 0.0119)} &\makecell{0.907916 \\ ($\pm$ 0.0001)}  & \makecell{\textbf{0.993936} \\ ($\pm$ 0.0013)} \\ 
\midrule 
Dataset-6 & \makecell{0.933144 \\ ($\pm$ 0.0002)} & \makecell{0.576750 \\ ($\pm$ 0.0111)} &\makecell{0.712136 \\ ($\pm$ 0.0420)}& \makecell{\textbf{0.986193} \\ ($\pm$ 0.0026)} \\  
\bottomrule
\end{tabular}
\caption{Performance (average precision) comparison of CHAD and baselines on trade datasets. CHAD shows a superior and consistent performance across all datasets. }
\label{tab:table_trade_results}
\vspace{-0.75em}
\end{table}

\subsection{Trade Data and Synthetic Anomalies}
% The shipment data utilized for experimental evaluation are subsets of the actual data used in the framework.
For experimental evaluation of our model, we use a subset of the real world trade data that is part of our proposed framework.
Six subsets are selected, where the training set is two months of data and the subsequent month is treated as the test set. 
Pre-processing steps includes removal of records with missing data and omitting entities which are rare  using a count threshold, following prior works~\cite{das2007detecting}.
The details of the subsets are shown in Table ~\ref{table:trade_data}
% =======================================

While generally evaluation of anomaly detection is performed using labelled data, considering one or more minority classes as anomalies, we do not have labels for trade data and thus generate synthetic anomalies following prior works~\cite{chen2016entity}.
For trade data, synthetic anomalous records are generated from the test set. 
Specifically, one categorical and one numeric attribute chosen at random are perturbed. 
For the categorical attribute, it is replaced with a randomly value(entity) belonging to the same field(domain). 
For the continuous valued attribute, a perturbation value ($p$) is added. If the current value is less than 0.5, $p$ is chosen from Uniform(0.25,0.75) and if the current value is greater than 0.5 $p$ is chosen from Uniform(-0.25,-0.75).
This approach is adopted to ensure that the records have unexpected patterns, characterized by the pair of perturbed attributes. Additionally anomalies generated should be not be trivial and easy to detect, for a proper evaluation of the model. This approach however is distinct from the way we generate negative samples, thus resulting in a fair comparison.
\subsection{Experimental Results on Trade Data}
We compare our model's performance against top-3 best performing models obtained from the evaluation above.
The model architecture is kept same, with similar values of dropout and a three layered autoencoder.
For all the autoencoder based approaches, same network architecture is used, specifically 3 layered fully connected network with dropout to obtain the latent representation.
The same hyperparameter settings for a particular model are used. The percentage of anomalies is set to $20\%$ instead of a balanced distribution, we use 5 sets of randomly sampled anomalies for each instance of model evaluation.
The results are shown in Table~\ref{tab:table_trade_results}. 
Our model is shown to outperform the baselines consistently. 
% More importantly, we observe that CHAD performs well for all datasets where as competing methods have a noticeable variability. 
This is significant for our purposes, since our usage scenario lacks any labelled data to perform hyperparameter tuning. 
% We omit DAGMM results due to the observed inconsistency of model performance in subsequent runs. We expand upon this in the next section.
% The metric chosen for comparison is \textit{average precision}, calculated as the area under precision-recall curve. 

\subsection{Comparing the Effect of Hyperparameters}\label{subsec:hyperparams}
Sensitive hyperparameters, which cannot be determined apriori without a validation set~\cite{oliver2018realistic}, have a significant effect on model performance.
We empirically show that competing models are highly sensitive to hyper-parameter values. The experiments are performed on the KDDCup-99 dataset where all models perform well.
Case in point, for OCSVM, although default value of $\nu$ is 0.5 is used in popular implementations and in the original work, we find setting it to 0.1 obtains best results. 
Further with a smaller number of anomalies to be detected, the hyperparameter setting affects performance more drastically as shown in Figure~\ref{fig:hyperparams}(a).
In case of deepSVDD authors recommend $\nu$ between 0.01 and 0.1, but the model performance varies significantly between those bounds, and beyond that. 
Moreover while the one-class objective is shown to consistently perform better in the original work, this is not always true as shown in Figure~\ref{fig:hyperparams}(a). 
In case of trade data soft-boundary objective works best for seepSVDD.
Similarly for DCN and DAGMM, the number of clusters has a significant effect on performance.
In training our model with negative samples, the count of negative samples per instance is the most significant hyper-parameter.
We train our model with varying number of negative samples per instance, and perform evaluation for all the data sets as shown shown in Figure~\ref{fig:hyperparams}(b). 
For the latter three datasets model performance is found to increase slightly with number of negative samples, though lower number of samples does not adversely affect the model performance. 
Thus our model is robust to choice of hyperparameters.
% This can be attributed to both quality of negative samples and instance noise injection. 
% This is in line with results shown in Section~\ref{sec:instance_noise}.
% Thus while our model is robust to number of negative samples,  using between 8 and 16 negative samples should lead to good performance.

\subsection{Varying Anomaly Content}
% An inherent challenge in anomaly detection is class imbalance.
% Not only are anomalous instances few and far between, they are expected not to follow a specific pattern. 
Given anomalies can be rare, it is important to understand how our model performs with varying percentage of anomalies. 
For each benchmark data set, we vary the percentages of anomalies in the combined test set from $2\%$ to $10\%$, to understand how our model performs in each scenario. 
The results are reported in Table~\ref{tab:vary_anom}. 
As expected, it is a more challenging task to detect anomalies when they are rarer; and our model performance improves slightly as the percentage of anomalies increase.
However it is also evident that CHAD is effective in cases with very low anomaly percentages.
\begin{table}[tp]
    \begin{tabular}
    {lccccc}
    \toprule
    Data Set & $2\%$ & $4\%$ & $6\%$ & $8\%$ & $10\%$ \\
    \midrule
    KDDCup99 &  0.9665  & 0.9705  & 0.9740  & 0.9739  & 0.9757\\
    KDD-N    & 0.9941 & 0.9958 & 0.9964 & 0.9967 & 0.9965\\
    NB15     & 0.7472 & 0.7630 & 0.7747 & 0.7834 & 0.7887\\
    NSL-KDD  & 0.6498 & 0.7042 & 0.7254 & 0.7401 & 0.7499\\
    Gure-KDD & 0.6905 & 0.7232 & 0.7357 & 0.7481 & 0.7560 \\
    \bottomrule
    \end{tabular}
     \caption{Average Precision score of CHAD for detecting anomalies, with different percentages of anomalies in the test set. CHAD is able to effectively detect even a small percentage(ratio) of anomalies.}
    \label{tab:vary_anom}
\end{table}
% \vspace{-0.5em}
% =======================================

% =======================================

% =======================================

% =======================================

\section{Model Analysis}
\begin{figure}[tp!]
    \centering
     \includegraphics[width=0.8250\columnwidth]{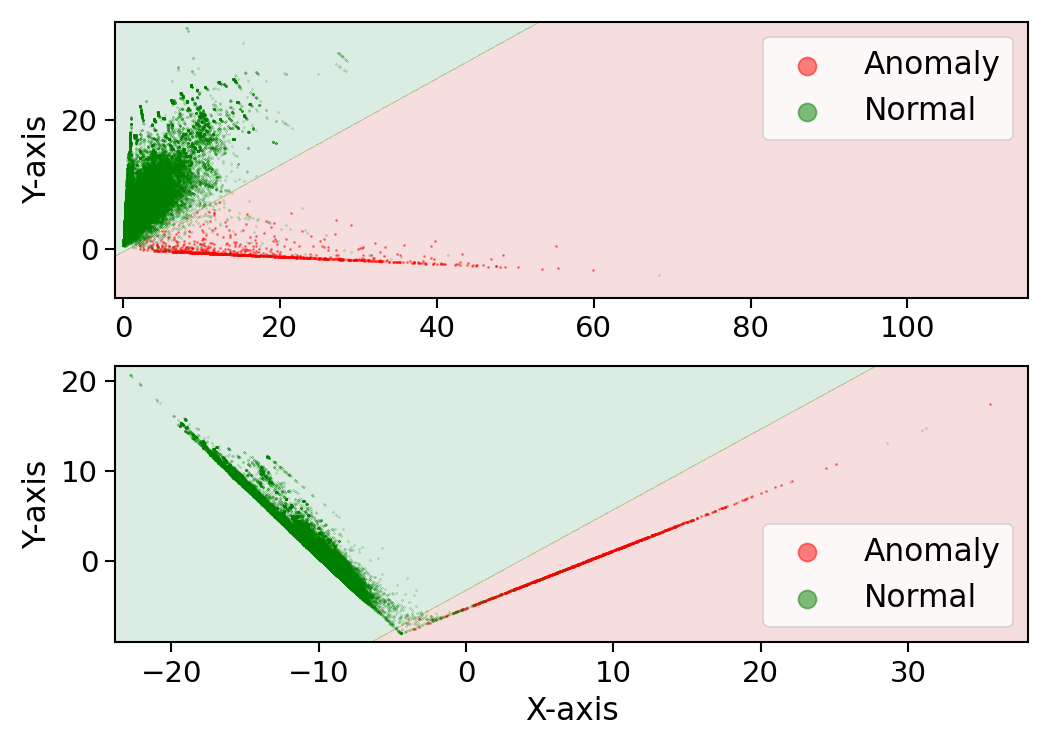}
    \caption{Visual representation of latent features for anomalies and nominal data points at different layers of the model. The top figure shows latent features obtained from the bottleneck layer of the asymmetric autoencoder, while the lower one shows features from penultimate layer of the density estimation network, both projected in a 2D space. With successive layers, we see the nominal and anomalous points are more distinctly separated.}
    \label{fig:viz_anom}
\end{figure}
 \vspace{-0.25em}
\subsection{Visualizing Anomalies}\label{subsec:anom_viz}
The utility of autoencoders in anomaly detection model architectures is to perform feature extraction and embed the data in a low dimensional space, where nominal points are clustered or are separable from anomalies.
To ascertain this, we perform visual analysis on feature vectors obtained from the lowest dimensional output of the autoencoder and the penultimate layer of the density estimation network as shown in Figure~\ref{fig:viz_anom}. We sample points from both the nominal data and anomalies, and perform dimensionality reduction to $R^2$ using SVD. For both the cases, we find that the nominal data is separable from the anomalies using a linear decision boundary in the transformed space.
This fortifies our hypothesis, and the model is able to separate nominal data instances from anomalies in the decision space.

\subsection{Importance of Secondary Noise}\label{sec:instance_noise}
\begin{figure}[tp]
\centering
\includegraphics[width=0.8\columnwidth]{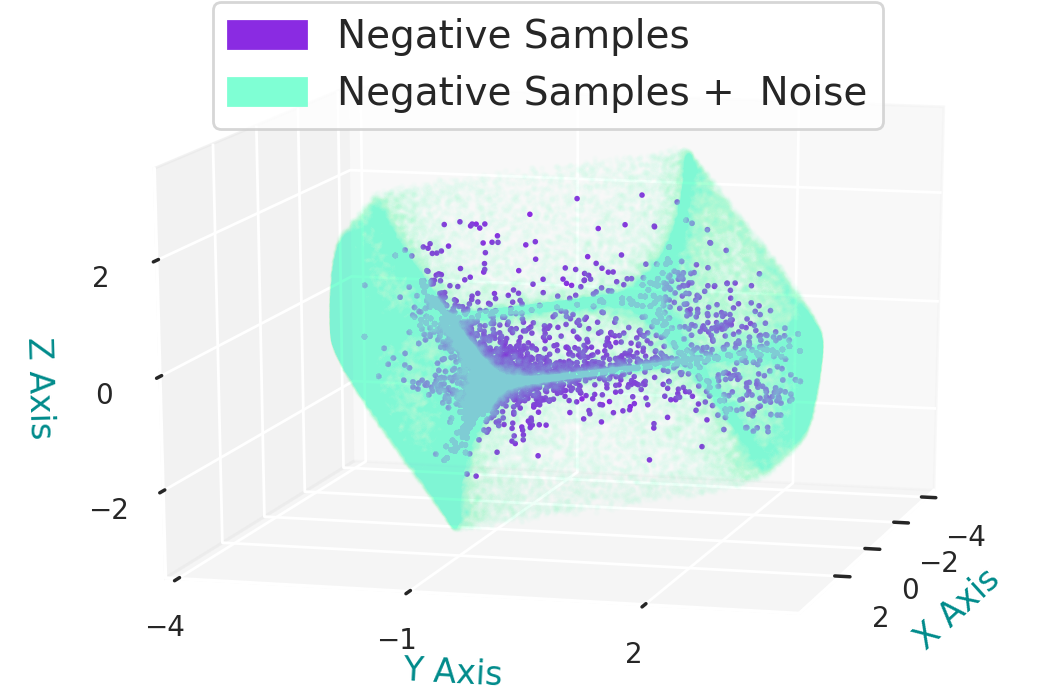}
\caption{Visualization of latent vectors of negative samples for KDDCup99 dataset, with and without injected noise. The addition of isotropic Gaussian noise results in a greater variation(spread) of negative samples across the latent space.}
\label{fig:noise_viz}
\end{figure}

We hypothesise that adding a secondary instance noise to negative samples at the latent space of the autoencoder helps in building a better model through a better approximation of the contrastive latent space, and our experiments confirms this. 
This isotropic Gaussian noise helps prevent over-fitting while training the model, which is crucial without validation sets in this unsupervised training scenario. 
This is evidenced by the fact that omitting the instance noise in some training runs, although the loss converges to a lower value the test time performance is lower. 
To better understand this, we visualize the negative samples generated for the dataset KDDCup99 with and without this noise, as shown in Figure~\ref{fig:noise_viz}.
It can be observed adding noise spreads out the latent representation of the negative samples over data space more evenly and across a larger region in the domain space of the nominal data. This effectively helps better estimate the space where nominal data does not occur.
The model thus gets trained with negative samples with a significantly greater variation and is more robust. 
Table~\ref{tab:noise_table} shows the improvement in average precision provided by the addition of this secondary noise, which is more pronounced  when only few negative samples per instance are used.
% \vspace{-0.5em}
\begin{table}[tp]
    \begin{tabular}
    {p{1cm}ccccc}
    \toprule
    Secondary noise  & KDD & KDD-N & NB15 & NSL-K  &  Gure-K  \\
    \midrule
    Yes       & \textbf{0.9723} &  0.9973 & 0.7874 & \textbf{0.7682} & \textbf{0.7662} \\
    No    & 0.9325 &  0.9974 & 0.7887 & 0.6983 & 0.6943 \\
    \bottomrule
    \end{tabular}
    \caption{Comparison of average precision scores when CHAD is trained with and without the secondary instance noise for negative samples. The secondary noise leads to significant overall improvement in model performance.}
    \label{tab:noise_table}
    % \vspace{-1.0em}
\end{table}

\section{Future Directions and Conclusion}
The anomaly detection method proposed in this work is part of a larger effort to build a framework for the real world task of detecting suspicious timber shipments.
Although prior methods have been shown to be effective on benchmark datasets, there remain several shortcomings that impede their application in real world systems with heterogeneous tabular data.
% The approach presented in this work tries to address some of the challenges of existing models in a systematic fashion.
We show our model CHAD can generalize to different data distributions and is robust to changes in hyper-parameters, and performs favorably against competing baselines. 
Our model has the potential to be applicable beyond timber shipment records, and could be applied to shipment records for any globally traded commodity, as well as be applicable to any heterogeneous tabular data occurring in other domains, as demonstrated by our comprehensive evaluation.

There are however further research questions we would like to address.
% This research being part of a larger project, our aim is to utilize the proposed approach as part of an overall framework with
% a human-in-the-loop system, that uses our model is a next step in that direction.
First relates to the issue of relevancy of anomalies, since only a subset of anomalies are relevant with respect to the application scenario. 
Improving relevancy of anomalies through active learning or a human-in-the-loop is a logical research direction that follows.
The second research question is to find ways of presenting the user with possible explanations as to why a record is judged anomalous. Explainability~\cite{lipton2018mythos} can be a powerful tool in garnering deeper understanding and trust of end users.
While we can visualize anomalies in latent space as shown in Section~\ref{subsec:anom_viz}, it does not provide a succinct reasoning. 
These challenges are applicable to our work in building the framework to detect suspicious timber shipments.
% Such an anomaly detection system can highlight potentially illegal or fraudulent trade records. 
% This can be potentially utilized for our next steps.
% A challenge that is faced by anomaly detection systems is that of relevancy, where a subset of anomalies are of interest to the user.
% This is a valid concern for our scenario as well.
% Since anomalies could be potentially used as starting point for further action --- false positives pose a problem.
% A human-in-the-loop based approach or active anomaly detection can help alleviate this issue.
% Additionally our model does not provide explanations towards why a record is judged anomalous. 
% Moving forward we would like also to explore the angle of fairness in anomaly detection.
Further research questions pertaining to temporal aspects of trade data and understanding feature importance are also of consequence for our application.
Therefore, there are multiple research directions as part of building our framework as well more general problems that stem from this work.
These questions motivate multiple continuing future research directions for us.
% \vspace{-0.5em}

\textbf{Acknowledgments}
This paper is based on work supported by
the NSF (DGE-1545362).

\bibliographystyle{ACM-Reference-Format}
\bibliography{references}
\end{document}